\documentclass[conference]{IEEEtran}
\IEEEoverridecommandlockouts
\usepackage{cite}
\usepackage{amsmath,amssymb,amsfonts}
\usepackage{algorithmic}
\usepackage{graphicx}
\usepackage{textcomp}
 \usepackage{tabularray}
 \usepackage[small]{caption}
\usepackage{xcolor}
\usepackage{amsthm}
\usepackage{etoolbox}
\def\BibTeX{{\rm B\kern-.05em{\sc i\kern-.025em b}\kern-.08em
    T\kern-.1667em\lower.7ex\hbox{E}\kern-.125emX}}
\usepackage{tikz}
\newtheorem{example}{Example}

\newtheorem{definition}{Definition}

\AfterEndEnvironment{definition}{\noindent\ignorespaces}
\begin{document}

\title{\huge ArgMed-Agents: Explainable Clinical Decision Reasoning with LLM Disscusion via Argumentation Schemes}

\author{
	\IEEEauthorblockN{Shengxin Hong, Liang Xiao, Xin Zhang, Jianxia Chen,}
	\IEEEauthorblockA{Hubei University of Technology, Wuhan, China}
	\IEEEauthorblockA{lx@mail.hbut.edu.cn}
}

\maketitle

\begin{abstract}
 There are two main barriers to using large language models (LLMs) in clinical reasoning. Firstly, while LLMs exhibit significant promise in Natural Language Processing (NLP) tasks, their performance in complex reasoning and planning falls short of expectations. Secondly, LLMs use uninterpretable methods to make clinical decisions that are fundamentally different from the clinician's cognitive processes. This leads to user distrust. In this paper, we present a multi-agent framework called ArgMed-Agents, which aims to enable LLM-based agents to make explainable clinical decision reasoning through interaction. ArgMed-Agents performs self-argumentation iterations via Argumentation Scheme for Clinical Discussion (a reasoning mechanism for modeling cognitive processes in clinical reasoning), and then constructs the argumentation process as a directed graph representing conflicting relationships. Ultimately, use symbolic solver to identify a series of rational and coherent arguments to support decision. We construct a formal model of ArgMed-Agents and present conjectures for theoretical guarantees. ArgMed-Agents enables LLMs to mimic the process of clinical argumentative reasoning by generating explanations of reasoning in a self-directed manner. The setup experiments show that ArgMed-Agents not only improves accuracy in complex clinical decision reasoning problems compared to other prompt methods, but more importantly, it provides users with decision explanations that increase their confidence.
\end{abstract}


\section{Introduction}
Large Language Models (LLMs) \cite{openai2023gpt4} have received a lot of attention for their human-like performance in a variety of domains. In the medical field especially, preliminary studies have shown that LLMs can be used as clinical assistants for tasks such as writing clinical texts \cite{nayak2023comparison}, providing biomedical knowledge \cite{singhal2022large} and drafting responses to patients' questions \cite{10.1001/jamainternmed.2023.1838}. However, the following barriers to building an LLM-based clinical decision-making system still exist: (i) LLMs still struggle to provide secure, stable answers when faced with highly complex clinical reasoning tasks \cite{pal2023medhalt}. (ii) There is a perception that LLMs use unexplainable methods to arrive at clinical decisions (known as black boxes), which may have led to user distrust \cite{eigner2024determinants}.

To address these barriers, exploring the capabilities of LLMs in argumentative reasoning is a promising direction. Argumentation is a means of conveying a compelling point of view that can increase user acceptance of a position. Its considered a fundamental requirement for building Human-Centric AI \cite{10.3389/frai.2022.955579}. As computational argumentation has become a growing area of research in Natural Language Processing (NLP) \cite{dietz2021computational}, researchers have begun to apply argumentation to a wide range of clinical reasoning applications, including analysis of clinical discussions \cite{QASSAS2015282}, clinical decision making \cite{10386040,KR2020-84,9dca3dfdee564220a341a3ba5d183a1c}, address clinical conflicting \cite{ee0bf59a8ab04a2fa1bf330ae9c379cc}. In the recent past, some work has assessed the ability of LLMs in argumentation reasoning \cite{chen2023exploring,castagna2024computational} or non-monotonic reasoning \cite{xiu-etal-2022-logicnmr}. Although LLMs show some potential for computational argumentation, more results show that LLMs perform poorly in logical reasoning tasks \cite{xie2024travelplanner}, and better ways to utilise LLMs for non-monotonic reasoning tasks need to be explored.

Meanwhile, LLM as agent studies have been surprisingly successful \cite{zhang2023cumulative,wang2023survey,shi2024ehragent}. These methods use LLMs as computational engines for autonomous agents, and optimise the reasoning, planning capabilities of LLMs through external tools (e.g. symbolic solvers, APIs, retrieval tools, etc.)  \cite{pan2023logiclm,shi2024ehragent}, multi-agent interactions \cite{tang2024medagents} and novel algorithmic frameworks \cite{gandhi2023strategic}. Through this design, LLMs agents can interact with the environment and generate action plans through intermediate reasoning steps that can be executed sequentially to obtain an effective solution.

Motivated by these concepts, we present ArgMed-Agents, a multi-agent framework designed for explainable clinical decision reasoning. We formalised the cognitive process of clinical discussion using an argumentation scheme for clinical discussion (ASCD) as a prompt strategy for interactive reasoning by LLM agents. There are three types of agents in ArgMed-Agents: the Generator, the Verifier, and the Reasoner. the Generator generates arguments to support clinical decisions based on the argumentation scheme; the Verifier checks the arguments for legitimacy based on the critical question, and if not legitimate, it asks the Generator to generate attack arguments; Reasoner is a LLM agent with symbolic solver that identifies reasonable, non-contradictory arguments in the resulting directed argumentation graph as decision support.

In our method, we do not expect every proposed argument or detection of Generator or Verifier to be correct, instead we consider their generation as a assumption. The LLM agents are induced to recursively iterate in a self-argumentative manner through the prompt strategy , while the newly proposed assumptions always contradict the old ones, and eventually the Reasoner eliminates unreasonable assumptions and identifies coherent arguments, leading to consistent reasoning results. ArgMed-Agents enables LLMs to explain its own outputs in terms of self-cognitive profiling by modelling their own generation as a prompt for question recursion.

Our experiment was divided into two parts, evaluating accuracy and explainability of ArgMed-Agents clinical reasoning, respectively. We conducted experiments on two datasets, including MedQA \cite{jin2020disease} and PubMedQA \cite{jin2019pubmedqa}. To better align with real-world application scenarios, our study focused on a zero-shot setting. The results show that ArgMed-Agents achieves better performance in both accuracy and explainability compared to direct generation and Chain of Thought (CoT).

In summary, we take following contribution:
\begin{itemize}
	\item We present a novel multi-LLM agent framework called ArgMed-Agents for complex clinical reasoning tasks.
	\item Our approach effectively combines argumentation frameworks and cognitive clinical medicine. Arguments are presented through LLM simulations of clinical discussions and reasoning results are solved through formal computational models, which avoids the cumulative errors of LLM logical reasoning and improves the safety of clinical reasoning.
	\item We propose the conjecture about ArgMed-Agents' theoretical guarantees now, which is that identify a reasoning error in the LLM when ArgMed-Agents considers all decisions unacceptable. This conjecture serves as a means for us to identify boundaries in the capabilities of large language models (e.g., hallucinatory phenomena, knowledge conflicts), which are important for clinical decision support. If the LLM makes a serious error but humans do not detect it in time, it may cause irreparable damage to the patient.
	\item We We perform an extensive evaluation of ArgMed-Agents, which demonstrate the accuracy, explainability and safety benefits in clinical decision support of our method.
\end{itemize}

\section{Preliminaries}

\subsection{Argumentation Scheme for Clinical Discussion}

Our approach utilises the notion of computational argumentation for to support reasoning. Abstract Argumentation (AA) \cite{DUNG1995321} are pair $\langle \mathcal{A},\mathcal{R} \rangle$ composed of two components: a set of abstract arguments $\mathcal{A}$, and a binary attack relation $\mathcal{R}$. Given an AA framework $\langle \mathcal{A},\mathcal{R} \rangle$, $a,b\in\mathcal{A}$ and $(a,b)\in\mathcal{R}$ indicates that $a$ attacks $b$. In this framework, there exist $a\in\mathcal{A}$ is acceptable if and only if:

\begin{itemize}
	\item not exist $ b\in\mathcal{A}$ such that $(b,a)\in \mathcal{R}$;
	\item there exist $b, c\in \mathcal{A}$ such that $(b,a)\in\mathcal{R}$ and $(c,b)\in \mathcal{R}$;
\end{itemize}

On top of this, numerous studies \cite{ORBi-e291a689-dab1-4c31-a12b-d523f431f7a3,9dca3dfdee564220a341a3ba5d183a1c,QASSAS2015282} have explored the application of argumentation in the clinical domain. In this section, we provide a summary of these endeavors, focusing on Argumentation Schemes for Clinical Discussion (ASCD). The concept of Argumentation Scheme (AS) originated within the domain of informal logic, stemming from the seminal works of \cite{Walton2008-WALAS,walton1996argumentation}. Argumentation Scheme serves as a semi-formalized framework for capturing and analyzing human reasoning patterns. Formally defined as $AS = \langle P, c, V\rangle$, it comprises a collection of premises ($P$), a conclusion ($c$) substantiated by these premises, and variables ($V$) inherent within the premises ($P.V$) or the conclusion ($c.V$). A pivotal aspect of the argumentation scheme is the delineation of Critical Questions (CQs) pertinent to AS. Failure to address them prompts a challenge to both the premises and the conclusion posited by the scheme. Consequently, the role of CQs is to instigate argument generation; when an AS is contested, it engenders the formulation of a counter-argument in response to the initial AS. This iterative process culminates in the construction of an attack argument graph, facilitating a nuanced understanding of argumentative dynamics.

\begin{figure}[htbp]
	\centering\includegraphics[width=0.35\textwidth,height=0.35\textwidth]{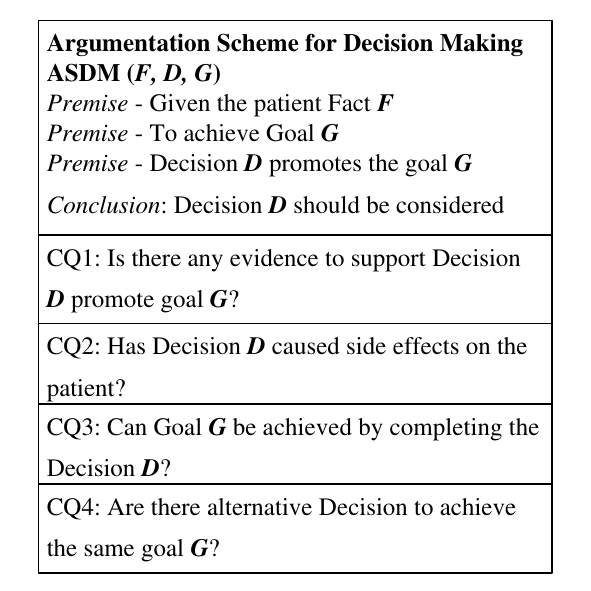} %插入文件名为fig1的图片，可以加文件后缀，如果是eps格式也可以不加。  
	% width=14cm 表示设置图片fig1的宽度为14cm
	\caption{An example of Argumentation Scheme for Decision Making }       %对图进行说明 
	\label{fig1}   
\end{figure}

ASCD encapsulates a variety of argumentation schemes that analyse the clinical discussion and reasoning process, such as Argumentation Scheme for Decision Making (ASDM), Argumentation Scheme for Side Effects (ASSE) and  Argumentation Scheme for Danger Apeal (ASDA). Figure \ref{fig1} illustrates an example of ASDM.

\section{Method}
In this section, we propose a multi-agent framework called \textit{ArgMed-Agents}, which supports the seamless integration of prompt strategy designed based on argumentation scheme for clinical discussion into agent interactions. Our approach  enhances LLMs to be able to perform explainable clinical decision reasoning without the need for expert involvement in knowledge encoding.

\subsection{ArgMed-Agents: a Multi-LLM-Agents Framework}

\begin{figure*}[htbp]
	\centering\includegraphics[width=1\textwidth,height=0.6\textwidth]{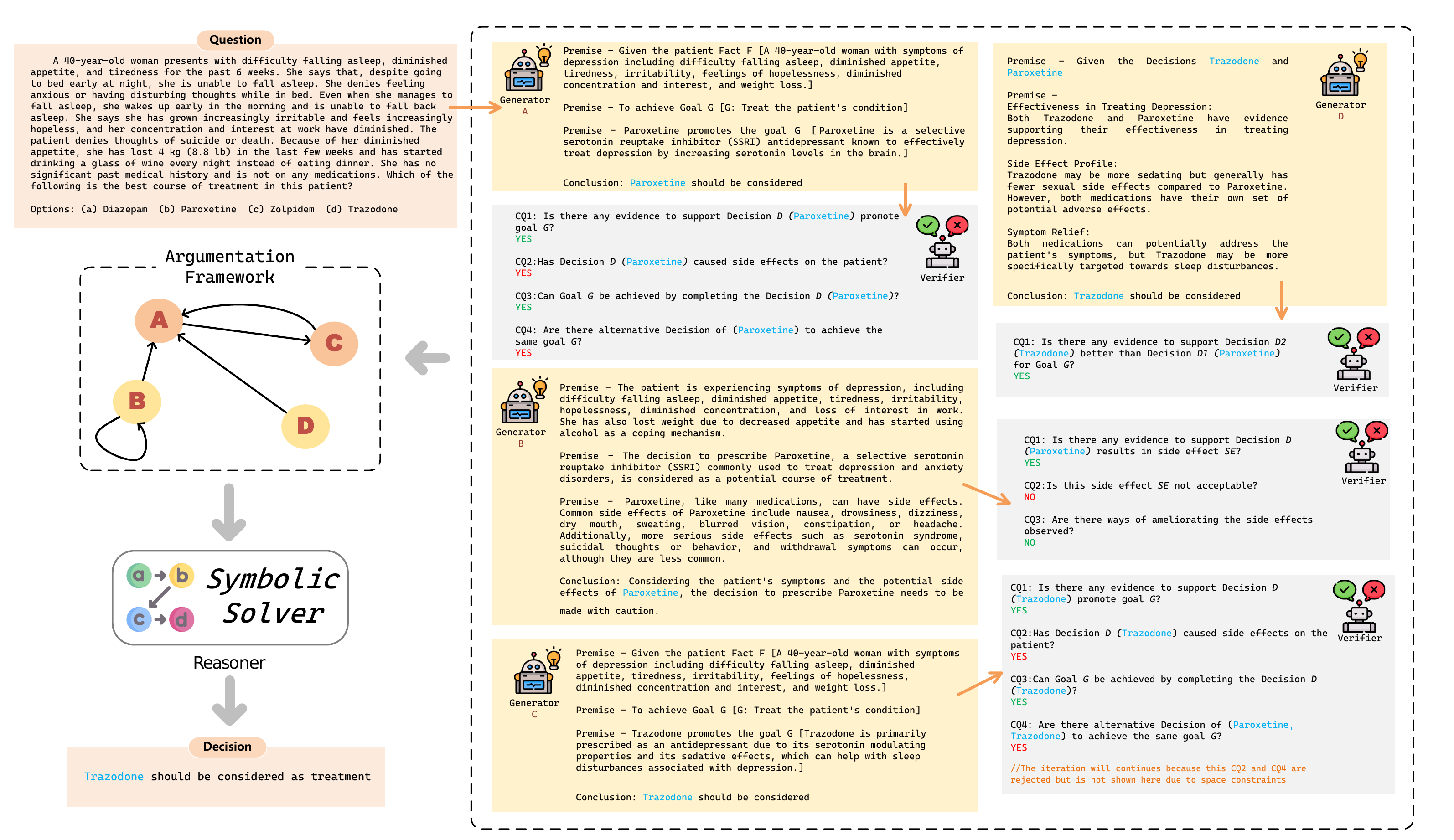} %插入文件名为fig1的图片，可以加文件后缀，如果是eps格式也可以不加。  
	% width=14cm 表示设置图片fig1的宽度为14cm
	\caption{An example from the MedQA USMLE dataset, with the entire process of ArgMed-Agents reasoning about the clinical problem. Notably, the letters in the argumentation framework correspond to the serial numbers of the four generators on the right, representing the premises and conclusion generated by that generator. In the argumentation framework, the red nodes ($A$ and $C$) represent arguments in support of the decision and the yellow nodes ($B$ and $D$) represent arguments in support of the beliefs.}       %对图进行说明 
	\label{fig2}   
\end{figure*}

ArgMed-Agents framework includes three distinct types of LLMs agent:
\begin{itemize}
	\item Generator(s): Generate arguments based on the current situation.
	\item Verifier: Whenever a new argument is generated, the verifier checks the accuracy of the argument by asking and answering CQs in a self-questioning manner. When the CQ validation is rejected, the reason why the CQ was rejected is returned to the generator, which proposes a new argument based on that CQ. This process is iterated until no more arguments are generated.
	\item Reasoner: It is an LLM agent equipped with a symbolic solver for computational argumentation. The reasoner records the complete process of iteration and identifies semantic relationships between arguments (attack or support). At the end of the iteration, all arguments form a complete abstract argumentation framework. The reasoner searches for a set of coherent and plausible arguments through the symbolic solver as a support for decision making.
\end{itemize}

In ArgMed-Agents, it do not expect single generation or single verification to be correct. ArgMed-Agents uses generation as a assumption to recursively prompt the model with critical questions, identifying conflicting and erroneous arguments in an iterative process. Ultimately, such mutually attacking arguments are converted into a formal framework that solves for a subset of reasonably coherent arguments via Reasoner.  Figure \ref{fig2} depicts how multi-agents interacts to simulate clinical discussion. In the example of Figure \ref{fig2}, the generator first proposes Proxetine as a therapeutic drug. However, after several rounds of discussion, ArgMed-Agents raised arguments about Proxetine's side effects, and eventually settled on the more effective therapeutic drug as Trazodone.
 
 The theoretical motivation for our method stems from non-monotonic logic, logical intuition and cognitive clinical. Studies have shown that LLM performs reasonably well for simple reasoning, single-step reasoning problems, however, as the number of reasoning steps rises, the rate of correct reasoning decreases in a catastrophic manner \cite{creswell2022selectioninference}. Thus, we consider that LLM is primed with logical intuition, but lacks true logical reasoning ability. 
 
 In light of this, ArgMed-Agents framework guides LLM in a recursive form to generate a series of  casual reasoning steps as a tentative conclusion, with any further evidence withdrawing their conclusion. The process ultimately leads to a directed graph representing the disputed relationships. Moreover, our method performs graph inference with the help of symbolic solvers, which avoids the cumulative error of LLM inference. LLM can be viewed as a large knowledge base containing a large amount of conflicting knowledge. ArgMed-Agents provides a formal method for discovering conflicts and solving for consistent reasoning results, which facilitates the LLM's ability to gradually come up with new knowledge to use in revising its own initial conclusion.
 
 Within the ArgMed-Agents framework, agents endowed with various LLM roles engage in collaborative interactions through a formalized process that simulate clinical discussion. This design not only facilitates LLMs' engagement in clinical reasoning through a critical-thinking approach, aligning their cognitive processes with those of clinicians to enhance decision explainability, but also enables the effective extraction and highlighting of implicit knowledge within LLMs that is not easily accessible through traditional prompts.
 
\subsection{Formal Computational Models \& Theorems}
In this section, we describe how ArgMed-Agents performs formal reasoning and explains decisions. First, we define the ArgMed-Agents interaction model

\begin{definition}[ArgMed-Agents Interaction] 
	In ArgMed-Agents, interaction is a series of dialogues $\mathcal{D}=\{s_1,...,s_n|n > 1\}$  involving generator
	agent $g$ and verifier agent $v$, and the following conditions hold:
	\begin{itemize}
		\item $s_i=\langle g,a \rangle$ is an arguments presented by generator $g$.
		\item $s_k=\langle v,cq \rangle$ is an critical question $cq\in CQs$ presented by verifier $v$.
		\item If there exist $s_i$ and $s_{i+2}$ from the generator then $s_{i+1}$ from the verifier such that $s_{i+2}$ attack $s_i$.
	\end{itemize}
\end{definition}

In ArgMed-Agents, we define the argument is presented by the generator and the verifier constructs an attack relation between the argument. Specifically, when the verifier rejects an argument, it leads to a new argument and this argument attacks the original argument; when the verifier accepts an argument, the round of dialogue ends. In this way, the reasoner can connect the semantic relations of all the arguments and construct an argumentation framework. Next, we define the argumentation framework for ArgMed-Agents:
\begin{definition}[Argumentation in ArgMed-Agents]
	An Argumentation Framework AF for ArgMed-Agents is a pair $\langle \mathcal{A},\mathcal{R} \rangle$ such that:
	\begin{itemize}
		\item $\mathcal{A}$ is arguments set constructed by ${s_i}\subseteq\mathcal{D}$ from generator and $\mathcal{R}$ is a set of binary attack relation.
		\item $\mathcal{A}=Args_d(\mathcal{A}) \cup Args_b(\mathcal{A})$ such that arguments in argument set $\mathcal{A}$ are distinguished between two types of arguments: arguments in support of decisions $Args_d(\mathcal{A})$ and arguments in support of beliefs $Args_b(\mathcal{A})$.
	\end{itemize}
\label{d2}
\end{definition}
We divided ArgMed-Agents' arguments into two types: arguments in support of decisions $Args_d(\mathcal{A})$ and arguments in support of beliefs $Args_b(\mathcal{A})$. These two types of arguments play different roles, $Args_d(\mathcal{A})$ build on beliefs and goals and try to justify choices, while $Args_b(\mathcal{A})$ always try to undermine decision arguments. In addition, we made the following setup for both arguments: First, arguments in support of different decisions are in conflict with each other; Second, arguments in support of decisions are not allowed to attack arguments in support of beliefs. Formalized as:

\begin{definition}
	Given an AF for ArgMed-Agents $\langle \mathcal{A},\mathcal{R} \rangle$, $\mathcal{R}\subseteq Args_d(\mathcal{A}) \times Args_b(\mathcal{A})$ such that:
	\begin{itemize}
		\item For all $a_1,a_2$ $\in Args_d(\mathcal{A})$ such that $a_1 \neq a_2,$ $(a_1,a_2), (a_2,a_1)\in \mathcal{R}.$
		\item Not exists $(arg_1,arg_2)$ $\in \mathcal{R}$ such that $arg_1\in Args_d(\mathcal{A})$ and $arg_2\in Args_b(\mathcal{A})$.
	\end{itemize}
\end{definition}
Definition \ref{4} describes how ArgMed-Agents identifies optional decisions and the concept of an explanation set for a decision.
\begin{definition}
	Given an AF for ArgMed-Agents $\langle \mathcal{A},\mathcal{R} \rangle$, if and only if a decision $d\in Args_d(\mathcal{A})$ is acceptable then this decision is optional decision. A set of decision support explanation $E$ include:
	\begin{itemize}
		\item There exist an decision $d=Args_d(E)$ is optional decision.
		\item All arguments $a\in E$ is acceptable and confict-free.
		\item Among the sets that satisfy the above two conditions, $E$ is the set that contains maximal elements.
	\end{itemize}
\label{4}
\end{definition}
Next, we use an example from MedQA to illustrate how Definition \ref{4} is applied in ArgMed-Agents.
\begin{example}[From MedQA]
	An 18-year-old woman presents with recurrent headaches. The pain is usually unilateral, pulsatile in character, exacerbated by light and noise, and usually lasts for a few hours to a full day. The pain is sometimes triggered by eating chocolates. These headaches disturb her daily routine activities. The physical examination was within normal limits. She also has essential tremors. Which drug is suitable in her case for the prevention of headaches?
\end{example}
\noindent After several rounds of discussion, ArgMed-Agents presented five arguments, including three decisions about treatments and two pieces of evidence about side effect, and they formed the following Argumentation framework:
\begin{figure}[htbp]
	\includegraphics[width=0.5\textwidth,height=0.16\textwidth]{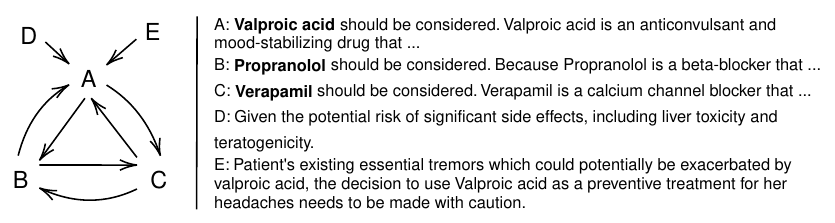} %插入文件名为fig1的图片，可以加文件后缀，如果是eps格式也可以不加。  
	% width=14cm 表示设置图片fig1的宽度为14cm
\end{figure}

\noindent Based on the definitions, we can obtain the following reasoning results:
\begin{itemize}
	\item Optional decisions: $B$ (Propranolol) or $C$ (Verapamil);
	\item Explaination set $E=\{\{B,D,E\},\{C,D,E\}\}$;
\end{itemize}
\noindent Notably, although two decisions are optional, we define the exclusivity of the decisions so that they do not appear in the decision support at the same time (this avoids duplicate medication). 

With this formal model, we present a conjecture about ArgMed-Agents' theoretical guarantee:

\begin{quote}
	{Given an AF for ArgMed-Agents $\langle \mathcal{A},\mathcal{R} \rangle$ and $E$ is its decisions support. There are errors in the clinical reasoning of ArgMed-Agents in this case iff $Args_d(E)=\varnothing$.} 
\end{quote}
Clinical reasoning errors are discussed in \cite{tang2024medagents}. They consider that most clinical reasoning errors in LLMs are due to confusion about domain knowledge. On the other hand, research by \cite{singhal2022large} points out that LLMs may produce compelling misinformation about medical treatment, so it is crucial to recognize this illusion, which is difficult for humans to detect. 
For this purpose, We conjecture that the relevant mechanisms for identifying clinical reasoning errors in ArgMed-Agents are as follows: When any decision in the $AF$ for ArgMed-Agents is not accepted, we consider there are errors in the reasoning by LLM and LLM's knowledge reserves are insufficient to address this issues. This mechanism assists ArgMed-Agents in identifying the capability boundaries of LLMs which helps to avoid the risks associated with adopting erroneous decisions and achieving more robust and safe clinical reasoning. We base this conjecture on the following:
ArgMed-Agents bootstrap the internal implicit knowledge of LLMs by iterating over them, which may be conflicting (possibly due to dirty data or timing conflicts). When there is only a small amount of conflicting knowledge, ArgMed-Agents can reason out the correct decision by non-monotonic reasoning; When there is a large amount of conflicting knowledge, the argumentation framework will be very large (containing erroneous arguments generated by hallucinatory phenomena), and the logical errors in this may lead ArgMed-Agents to be unable to reason out any correct decisions. We therefore consider this situation as the boundary of the LLM's capabilities.

\section{Experiments}
In this section, we demonstrate the potential of ArgMed-Agents in clinical decision reasoning by evaluating the accuracy and explainability.

\subsection{Settings}
We implemented different types of agents in ArgMed-Agents using the APIs GPT-3.5-turbo and GPT-4 provided by OpenAI \cite{openai2023gpt4}, and according to our setup, the LLM Agents are implemented with the same LLM and different few-shot prompts. Each agent is configured with specific parameters: the temperature is set to 0.0, as well as a dialogue limit of 8 and the maximum number of decisions allowed for ArgMed-Agents to generate limit of 4 (because the MedQA dataset is multiple choice with only four options), which is to prevent the agents from getting into loops with each other. On the other hand, we using python3 to implemente a symbolic solver of our formal model.
\subsection{Datasets}
The following two datasets were used to assess the accuracy and explainability of ArgMed-Agents clinical reasoning:
\begin{itemize}
	\item MedQA \cite{jin2020disease}:Answering multiple-choice questions derived from the United States Medical License Exams (USMLE). This dataset is sourced from official medical board exams and encompasses questions in English, simplified Chinese, and traditional Chinese. The total question counts for each language are 12,723, 34,251, and 14,123, respectively.
	\item PubMedQA \cite{jin2019pubmedqa}: A biomedical question and answer (QA) dataset collected from PubMed abstracts. The task of PubMedQA is to answer yes/no/maybe research questions using the corresponding abstracts (e.g., Do preoperative statins reduce atrial fibrillation after coronary artery bypass graft surgery).
\end{itemize}
\subsection{Accuracy}
In accuracy evaluation, we rigorously curated datasets by randomly selecting 300 examples from each, ensuring a balanced representation for robust analysis. To establish a baseline for comparison, we employed GPT direct generation alongside Chain of Thought (CoT) as outlined in \cite{wei2023chainofthought}. Notably, our primary focus centered on evaluating the efficacy of ArgMed-Agents within the domain of clinical decision reasoning. It is pertinent to highlight that while datasets such as MedQA and PubMedQA contained biomedical general knowledge quiz-type questions, we intentionally excluded this subset during the example selection process. This deliberate exclusion aimed to maintain the integrity of our evaluation specifically within the realm of clinical decision-making, thereby ensuring a targeted assessment of ArgMed-Agents' performance.

\begin{table}
	\centering
	\begin{tblr}{
			hline{1,8} = {-}{0.08em},
			hline{2,5} = {-}{},
		}
		Model         & Method        & MedQA & PubMedQA \\
		& Direct        & 52.7     & 68.4        \\
		GPT-3.5-turbo & CoT           & 48.0     & 71.5        \\
		& ArgMed-Agents & 62.1     & 78.3        \\
		& Direct        & 67.8     & 72.9        \\
		GPT-4         & CoT           & 71.4     & 77.2        \\
		& ArgMed-Agents & 83.3     & 81.6        
	\end{tblr}
	\caption{Results of the accuracy of various clinical reasoning methods on MedQA and PubMedQA datasets.}
	\label{t1}
\end{table}

Table \ref{t1} presents the accuracy outcomes for MedQA and PubMedQA, contrasting ArgMed-Agents with various baselines under direct generation and Chain of Thought (CoT) configurations. Notably, evaluations employing GPT-3.5-turbo and GPT-4 models revealed that our proposed ArgMed-Agents notably enhanced accuracy in clinical decision reasoning tasks compared to CoT and other baselines. This finding underscores the efficacy of ArgMed-Agents in augmenting the clinical reasoning capabilities of Language Model (LLM) systems. Intriguingly, through multiple experiment repetitions, we observed instances where integrating CoT resulted in an unexpected decline in performance. This phenomenon might stem from the propensity of LLMs to experience hallucinatory behaviors, exacerbated by CoT's potential to perpetuate such hallucinations indefinitely. In contrast, ArgMed-Agents corrects initial erroneous conclusions by means of iterative argumentation, effectively mitigates this issue.

\subsection{Explainability}
The main focus of Explainable AI (XAI) is usually the reasoning behind decisions or predictions made by AI that become more understandable and transparent. In the context of XAI, Explainability is defined as follows \cite{ISOIEC:AWITS29119-11}:
\begin{itemize}
	\item[] \textit{...level of understanding how the AI-based system\\
		 came up with a given result.}
\end{itemize}

Based on this criterion, we define measures of LLM's explanability:
\begin{itemize}
	\item How far do the explanations given by LLM help users to predict LLM's decisions?
\end{itemize}

Our team has produced a fully functional knowledge-based clinical decision support system (CDSS) \cite{10385915} at an early stage that can be used to assist in treatment decisions for complex diseases such as cancer, neurological disorders, and infectious diseases. This CDSS consist of computer-interpretable guidelines in the form of Resource Description Framework (RDF) \cite{klyne2004resource} and the corresponding inference engine. Knowledge-based CDSS can be viewed as explainable systems.

With this, our experimental setup is as follows: We set the direct generation explanation and COT as the baseline and the knowledge-based CDSS as the benchmark of what we expect explainability of ArgMed-Agents to be close to. We documented the input (i.e., questions) and reasoning process for 100 examples in MedQA (e.g., COT's chain of reasoning, RDF inference nodes in knowledge-based CDSS and the complete dialogue between ArgMed-Agents and the corresponding argumentation framework). For evaluation, We fine-tuned an LLM to act as an evaluator, causing it to predict the corresponding decision based on the inference record. We considered the accuracy of the evaluator's predictions as an indicator of interpretability (i.e., how well the reasoning process helps the user understand the decision).
\begin{table}
	\centering
	\label{t2}
	\begin{tblr}{
			hline{1-2,5,8-9} = {-}{},
		}
		Model         & Method               & \textit{Pre.} \\
		& Direct               & 0.53           \\
		GPT-3.5-turbo & CoT                  & 0.59           \\
		& ArgMed-Agents        & 0.87           \\
		& Direct               & 0.68           \\
		GPT-4         & CoT                  & 0.73           \\
		& ArgMed-Agents        & 0.91           \\
		& Knowledge-based CDSS & 0.95           
	\end{tblr}
\caption{Predict accuracy (\textit{Pre.}) with reasoning record of different models and methods in 100 MedQA examples.}
\end{table}
The knowledge-based CDSS achieved the highest predictive accuracy (0.95) as the benchmark for explainability. Among the tested methods, ArgMed-Agents demonstrated significantly higher predictive accuracy compared to direct and CoT methods in both GPT-3.5-turbo and GPT-4 models. ArgMed-Agents with GPT-4 achieved 0.91 predictive accuracy, closely approaching the knowledge-based CDSS's level of explainability.

The study demonstrates that ArgMed-Agents significantly enhance the explainability of LLMs, enabling users to better understand and predict the models' decisions. This suggests that employing advanced argumentation frameworks can bridge the gap between black-box AI models and fully transparent, knowledge-based systems like the CDSS.

\subsection{Discussion}
Conjecture 1 posits that errors in clinical reasoning occur in ArgMed-Agents if the AF does not yield any acceptable decisions. To provide a preliminary proof, we analyzed instances where ArgMed-Agents failed to produce acceptable decisions and correlated these instances with the observed errors. Our data indicates that in 63\% of cases where ArgMed-Agents failed to yield an acceptable decision set, there were identifiable clinical reasoning errors. These errors were primarily due to the presence of extensive conflicting knowledge within the AF, which prevented the system from arriving at a coherent decision. Specifically, 76\% of these errors were linked to confilct knowledge, while 24\% were due to insufficient domain knowledge. 

Our analysis suggests that the capacity of ArgMed-Agents to navigate and resolve conflicts in medical knowledge is a critical determinant of its effectiveness. This is consistent with the assumption as follows: for a given input (e.g., a clinical reasoning problem), when only a small amount of conflicting knowledge exists in the LLM, ArgMed-Agents can reason out the correct treatment plan by vetoing out the erroneous arguments through further argumentation. When there is a large amount of inconsistent knowledge in the LLM, these conflicts lead to a complex system that prevents the LLM from maintaining logical consistency in its generation (i.e., the phenomenon of hallucination), thus hindering effective decision-making.

The advantage of ArgMed-Agents over similar existing techniques is that the conditions under which LLM reasoning fails (i.e. when faced with a large amount of conflicting knowledge or insufficient domain knowledge) can be recognised through formal arguments, allowing us to better understand and predict the limitations of LLM in clinical applications. This understanding is critical to mitigating the risks associated with poor decision-making in healthcare settings.

In addition to this, we analyse the traceability of ArgMed-Agents reasoning. We find that the reasoning process by which the CoT method arrives at an answer is usually logically incomplete, whereas the reasoning process of ArgMed-Agents is much more refined. Interestingly, we found that the reasoning of the knowledge-based CDSS on many examples can be regarded as a reasoning subgraph of ArgMed-Agents, probably because ArgMed-Agents traverses all the reasoning paths in a randomly generated manner, and thus the reasoning graph includes not only the reasoning paths of the correct decisions, but also the explanations of the incorrect decisions. This finding provides further evidence of the explainability of ArgMed-Agents. In many clinical decision cases, patients may want to understand not just why a particular decision was taken, but why another decision could not be adopted.

\section{Related Work}
\subsection{LLM-based Clinical Decision Support System}
Extensive research underscores the potential utility of Large Language Models (LLMs) in medical applications \cite{bao2023discmedllm,nori2023generalist,jin2023genegpt}. However, these models encounter challenges in making reliable decisions when faced with complex clinical scenarios that demand advanced medical expertise and robust reasoning skills \cite{singhal2022large}. Consequently, significant efforts have been initiated to augment the clinical reasoning capabilities of LLMs. \cite{tang2024medagents} introduces a Multi-disciplinary Collaboration (MC) framework employing LLM-based agents in simulated role-playing scenarios, facilitating collaborative, iterative discussions aimed at consensus building. Despite yielding promising outcomes, this approach struggles to formalize iterative results effectively to enhance the inference performance of LLMs using dedicated inference tools.

Another approach, proposed by \cite{savage2024diagnostic}, leverages diagnostic reasoning prompts to enhance clinical reasoning and interpretability in LLMs. However, this approach improves explainability through templates designed for specific diseases, whereas ArgMed-Agents is more automated. In addition to this, ArgMed-Agents provides explanations for why a decision would not have been chosen.

Additionally, research efforts such as those by \cite{singhal2023expertlevel,li2023llavamed} involve fine-tuning LLMs using extensive datasets sourced from medical and biomedical literature. In contrast to these approaches, our method focuses on exploiting latent medical knowledge inherently present within LLMs to enhance their reasoning abilities in a training-free setting.

\subsection{Logical Reasoning with LLMs}
An extensive body of research has been dedicated to utilizing symbolic systems to augment reasoning, which encompasses a variety of methodologies including code environments, knowledge graphs, and formal theorem provers \cite{pan2023logiclm,Pan_2024,wu2023autogen}. In the study conducted by Jung et al. \cite{jung2022maieutic}, reasoning is framed as a satisfiability problem of its logical relations through the application of inverse causal reasoning. This approach leverages SAT solvers to enhance consistent reasoning and thereby improves the reasoning capabilities of Large Language Models (LLMs). 

Zhang et al.'s work \cite{zhang2023cumulative} takes a different approach by employing cumulative reasoning to break down tasks into smaller, more manageable components, which simplifies the problem-solving process and boosts overall efficiency. In another study, Xiu et al. \cite{xiu-etal-2022-logicnmr} explore the non-monotonic reasoning capabilities of LLMs. However, despite the promising initial results demonstrated by LLMs, their performance is notably inadequate in terms of generalization and proof-based traceability, with a marked decline in effectiveness as the depth of reasoning increases. 

Consistent with our findings, recent studies have begun to investigate the potential for enhancing argumentative reasoning in LLMs \cite{chen2023exploring}. These efforts aim to bolster the argumentative reasoning abilities of LLMs, as evidenced by the work of Dewynter et al. \cite{dewynter2023i} and Castagna et al. \cite{castagna2024computational}, which strive to further develop this aspect of LLM performance.

\section{Conclusion}
In this work, We propose a novel medical multi-agent interaction framework called ArgMed-Agents, which inspires agents of different roles to simulate the process of clinical discussion, iterating through argumentation and critical questioning. Finally, Reasoner agent to identify a set of reasonable and coherent arguments in this framework as decision support and explaintion. The experimental results indicate that, compared to different baselines, using ArgMed-Agent for clinical decision-making reasoning achieves greater accuracy and provides inherent explanations for its inferences. In addition to this, the analyses in our discussion show that ArgMed-Agents are able to identify their own reasoning errors and ability boundaries to a large extent. This means that ArgMed-Agents provides safer decisions compared to other state-of-the-art methods. Our goal in this work is to provide healthcare professionals with powerful tools to enhance their decision-making process and ultimately improve patient outcomes.

Despite the success of ArgMed-Agents, there are still some limitations. We suggest the following future directions for the linmitations: (1) While we used abstract argumentation to formalise the results of the ArgMed-Agents clinical discussion, extension to more expressive logics (e.g., first-order logic, descriptive logic, modal logic, or probabilistic logic) would allow for more sophisticated reasoning and argument generation. (2) Adaptive approaches that tailor arguments to the needs and preferences of individual users can further improve the effectiveness and explainability of clinical decisions. Future work could explore interactive clinical decision support systems based on patient/physician feedback.

\bibliographystyle{IEEEtran}
\bibliography{myBibtex.bib}

\end{document}